%% file: coling2020.tex
\documentclass[11pt]{article}
\usepackage{coling2020}
\usepackage{times}
\usepackage{url}
\usepackage{latexsym}
\usepackage{longtable}
\usepackage{pgfplots}
\usepackage{amsmath}
\usepackage{tikz}
\usepackage{mathdots}
\usepackage{yhmath}
\usepackage{cancel}
\usepackage{color}
\usepackage{siunitx}
\usepackage{array}
\usepackage{multirow}
\usepackage{amssymb}
\usepackage{textcomp}
\usepackage{gensymb}
\usepackage{tabularx}
\usepackage{booktabs}
\usepackage{enumerate}
\usepackage{lipsum}
\usepackage{makecell}
\usepackage{subcaption}
\usepackage{verbatim}
\pgfdeclarelayer{background}% determine background layer
\pgfsetlayers{background,main}% order of layers
\pgfplotsset{compat=1.14}
\definecolor{bblue}{HTML}{4F81BD}
\definecolor{rred}{HTML}{C0504D}
\definecolor{ggreen}{HTML}{9BBB59}
\definecolor{ppurple}{HTML}{9F4C7C}
\pgfplotsset{
   compat=1.14,
   legend entry/.initial=,
   every axis plot post/.code={%
       \pgfkeysgetvalue{/pgfplots/legend entry}\tempValue
       \ifx\tempValue\empty
           \pgfkeysalso{/pgfplots/forget plot}%
       \else
           \expandafter\addlegendentry\expandafter{\tempValue}%
       \fi
   },
}
\newcolumntype{Y}{>{\hsize=.4\hsize}X}
\title{Beyond Leaderboards: A survey of methods for revealing\\weaknesses in Natural Language Inference data and models}

\author{Viktor Schlegel, Goran Nenadic {\normalfont and} Riza Batista-Navarro\\
  Department of Computer Science, University of Manchester \\
  Manchester, United Kingdom \\
  {\tt \{viktor.schlegel, gnenadic, riza.batista\}@manchester.ac.uk}
  }

\date{}

\begin{document}
\maketitle
\begin{abstract}
Recent years have seen a growing number of publications that analyse Natural Language Inference (NLI) datasets for superficial cues, whether they undermine the complexity of the tasks underlying those datasets and how they impact those models that are optimised and evaluated on this data.
This structured survey provides an overview of the evolving research area by categorising reported weaknesses in models and datasets and the methods proposed to reveal and alleviate those weaknesses for the English language. We summarise and discuss the findings and conclude with a set of recommendations for possible future research directions.
We hope it will be a useful resource for researchers who propose new datasets, to have a set of tools to assess the suitability and quality of their data to evaluate various phenomena of interest, as well as those who develop novel architectures, to further understand the implications of their improvements with respect to their model's acquired capabilities.
\end{abstract}

\section{Introduction}
%\cite[MNLI;][]{rajpurkar2016squad,rajpurkar2018know}
%\subsection{Motivation}
Research in areas that require natural language inference (NLI) over text, such as Recognizing Textual Entailment (RTE) \cite{Dagan2006TheChallenge} and Machine Reading Comprehension (MRC) is advancing at an unprecedented rate. On the one hand, novel architectures \cite{Vaswani2017} enable efficient unsupervised training on large corpora to obtain expressive contextualised word and sentence representations for a multitude of downstream NLP tasks \cite{Devlin2018}. On the other hand, large-scale datasets \cite{Bowman2015,rajpurkar2016squad,Williams2018} provide sufficient examples to optimise large neural models that are capable of outperforming the human baseline on multiple tasks \cite{Raffel2019ExploringTransformer,Lan2020ALBERT:Representations}.

%-- great performance in NLP tasks, thanks to architectures \cite{Vaswani2017} that allow efficient training of large-scale contextualised representations \cite{Devlin2018} 

%-- large-scale crowd-sourced datasets allow training of neural models on a lot of examples

%-- those factors result in models that even surpass human performance in NLP tasks 
%These findings prompted researchers to look beyond the simple one-dimensional performance metrics as they are typically reported by leader boards, in search for an alternative explanation for the seemingly superb performance. In fact, those neural models appear to exploit cues and spurious correlations in training data, such as lexical overlap between sentences \cite{mccoy2019right}, certain words \cite{Poliak2018} or sentence length \cite{gururangan2018annotation} that correlate with expected predictions. When evaluated on data in which those cues have been removed, their performance deteriorates significantly \cite{mccoy2019right,Niven2019ProbingArguments} showing that the models are in fact relying on them rather than learning to understand meaning or perform inference.

Recent work, however, has questioned the seemingly superb performance for some of the tasks. Specifically, training and evaluation data may contain exploitable superficial cues, such as syntactic constructs \cite{mccoy2019right}, specific words \cite{Poliak2018} or sentence length \cite{gururangan2018annotation} that are predictive of the expected output. After having been evaluated on data in which those cues have been removed, the performance of those models deteriorated significantly \cite{mccoy2019right,Niven2019}, showing that they are in fact relying on the existing cues rather than learning to understand meaning or perform inference.
%An example of systematic errors committed by such models and spurious correlations present in data is shown in Figure~\ref{fig:example}. 
In other words, those well-performing models tend to obtain optimal performance on a particular dataset, i.e. overfitting on it, rather than generalising for the underlying task. 
This issue, in fact, remains concealed, if a model is compared to a human baseline by means of a single number that reports the average score on a held-out test set, which is typically the case with contemporary benchmark leaderboards.
%This issue, in fact, cannot be revealed by measuring models by means of a single one-dimensional performance metric, as they are typically reported by a benchmark's leader board.

To reveal and overcome these issues mentioned above, a growing number of approaches has been proposed in the past. 
% They can be grouped into three subcategories, namely methods that:
% \begin{enumerate}[(a)]
%     \item reveal systematic issues with existing training and evaluation data, such as the spurious correlations mentioned above
%     \item investigate what inference and reasoning capabilities models optimised on these data acquire when evaluated on samples not drawn from the training distribution, either by providing a collection of different benchmarks \cite{Dua2019} or different tasks \cite{Wang2019SuperGLUE:Systems}, or by investigating a particular capability more closely, such as processing hypo- and hypernymy \cite{Glockner2018}.
%     \item propose architectural \cite{Sagawa2019} and training procedure \cite{Wang2018} improvements in order to achieve more robust generalisation beyond data drawn from the training distribution, i.e. forcing models to ignore the biases present in training data to a certain extent.
% \end{enumerate}
All those methods contribute towards a fine-grained understanding of whether the existing methodology actually evaluates the required inference capabilities, what existing models learn from available training data and, more importantly, which capabilities they still fail to acquire, thus providing targeted suggestions for future research. %to solve them or whether they are solvable through shortcuts such as lexical cues. 

To make sense of this growing body of literature and help researchers new to the field to navigate it, we present a structured survey of the recently proposed methods and report the trends, applications and findings. 
In the remainder of this paper, we first establish terminology, set the objectives and the scope of the survey and describe the data collection methodology. We then present a categorisation of the surveyed methods with their main findings, and finally discuss the arising trends and open research questions.

% language!
\begin{figure}[t]
    \centering
    \begin{minipage}[b]{0.32\columnwidth}
        %\begin{table}[t]
            \centering
            \begin{tabular}{lll}
                 %\hline
                 %\makecell[tl]{Heuristic} & \makecell[l]{Supporting\\ Cases} &\makecell[l]{Contradicting\\ Cases}\\
                 \makecell[tl]{Heuristic} & \makecell[l]{$E$} &\makecell[l]{$\neg E$}\\
                 \hline
                 Lex. Overlap & 2,158 &261\\
                 Subsequence & 1,274 & 72\\
                 Constituent & 1,004 & 58
                 %\hline
            \end{tabular}
        %\end{table}
        %\vspace{-2\baselineskip}
        \caption{Number of premise-hypothesis pairs in an RTE dataset following lexical patterns, spuriously skewed towards \emph{Entailment}
        %Lexical patterns in hypothesis and premise of an RTE dataset spuriously correlate with the label
        \cite{mccoy2019right}.}
        \label{fig:example-correlations}
    \end{minipage}%
    \quad
    \begin{minipage}[b]{0.65\columnwidth}
    \captionsetup{width=.95\linewidth}
    \centering
    \vspace{0.1\baselineskip}
    \begin{tabularx}{0.95\textwidth}{| X |}
        \hline
        \footnotesize
                 %\makecell[tl]{Heuristic} & \makecell[l]{Supporting\\ Cases} &\makecell[l]{Contradicting\\ Cases}\\
                 \textbf{Paragraph}: \emph{``[\ldots] The past record was held by {\color{green}John Elway}, who led the Broncos to victory in Super Bowl XXXIII at age 38 and is currently Denver’s Executive Vice President of Football Operations and General Manager. {\color{blue}Quarterback Jeff Dean had jersey number 37 in Champ Bowl XXXIV.''}}\par
                 \textbf{Question}: \emph{``What is the name of the quarterback who was 38 in Super Bowl XXXIII?''}\par
                 \textbf{Original model prediction}: {\color{green}John Elway}\par
                 \textbf{Model prediction after inserting a distracting sentence}: {\color{red}Jeff Dean}\\
                 
        \hline         
        \end{tabularx}
    %\vspace{-1.4\baselineskip}
    \caption{Models' over-stability towards common words in question and paragraph, revealed by adversarially inserting distracting sentences \cite{Jia2017}.}
    \label{fig:example-overstability}
    \end{minipage}
\end{figure}

% \begin{figure}[t]
%     \centering
% \begin{subfigure}[t]{0.5\textwidth}
%     \centering
%     %\include{figures/adversarial-evaluation}
%     \includegraphics[width=1\textwidth]{figures/example2}
% \end{subfigure}%
% \begin{subfigure}[t]{0.5\textwidth}
%     \centering
%     \includegraphics[width=1\textwidth]{figures/probability-mass}
% \end{subfigure}
% \caption{Left: Example for model over-stability towards keywords shared between question and a distracting adversary sentence \cite{Jia2017}; Right: spurious correlations between hypothesis length and expected label \cite{gururangan2018annotation}}.
% \label{fig:example}
% \end{figure}

% \begin{table}[!tb]% \end{table}

% \end{table}

% \centering
% \begin{tabularx}{1\columnwidth}{l  X  X}
% \hline
% Heuristic & Definition & Example \\
% \hline
% Lexical overlap & Assume that a premise entails all hypotheses constructed from words in the premise  &   The doctor was paid by the actor. \\
% \hline
% \end{tabularx}
% \caption{Think of some flashy example. This one is (in part) from \newcite{mccoy2019right}.}
% \label{table:selected-datasets}
%Research in NLP, particularly in the areas that require inference over unstructured text, such as Recognizing Textual Entailment (RTE) (?Dagan, 2006) and Machine Reading Comprehension (MRC) is advancing at an unprecedented rate. On the one hand, novel architectures \cite{Vaswani2017} enable efficient unsupervised training on large corpora to obtain expressive contextualised word and sentence repres

% \end{table}

\subsection{Terminology}

\paragraph{Tasks:} The task of \emph{Recognising Textual Entailment (RTE)} is to decide, for a pair of natural language sentences (premise and hypothesis), whether given the premise the hypothesis is true (\emph{Entailment}), false (\emph{Contradiction}) or whether the two sentences are unrelated (\emph{Neutral}) \cite{Dagan2013RecognizingApplications}.

We refer to the task of finding the correct answer to a question over a passage of text as \emph{Machine Reading Comprehension (MRC)}, also known as Question Answering (QA). Usual formulations of the task require models to select a span from the passage, select from a given set of alternatives or generate a free-form string \cite{Liu2019a}. 

In this paper, we use the term ``NLI'' in its broader sense, referring to the requirement to perform inference over natural language. Thus we expand the usual textual entailment-based definition to also include MRC, as answering a question can be framed as finding an answer that is entailed by the question and the provided context, and the tasks can be transformed vice versa \cite{Demszky2018TransformingDatasets}.
%\begin{itemize}
\paragraph{Model and Architecture:} We refer to the neural network architecture of a model as ``architecture'', e.g. BiDAF \cite{Seo2017}.
We refer to a (statistical) model of a certain architecture that was optimised on a given set of training data simply as ``model''. It is important to make this distinction, as an optimised model's systematic failures can either be traced to biases in the training data (and can potentially be different for a model optimised on different data) or affiliated with the model class (and exist for all models with the same architecture) \cite{Liu2019,Geiger2019}.

\paragraph{Spurious Correlations:} We call correlations between input data and the expected prediction as ``spurious'' if they are not indicative for the underlying task but rather an artefact of the data at hand (as illustrated in Figure~\ref{fig:example-correlations}). The exploitation of those correlations in order to produce the expected prediction is known as the ``Clever Hans Effect'', named after a horse that was believed to perform arithmetic tasks but was shown to react to subtle body language cues of the asking person \cite{Johnson1911}. 

\paragraph{Adversarial:} \newcite{Szegedy2014} define ``adversarial examples'' as (humanly) imperceptible perturbations to images that cause a significant drop in the prediction performance of neural models. Similarly for NLP, we refer to data as ``adversarial'' if it is designed to minimise prediction performance for a class of models, while not impacting the human baseline. Examples include appending irrelevant information \cite{Jia2017}, illustrated in Figure~\ref{fig:example-overstability}, or paraphrasing \cite{Ribeiro2019}.

\paragraph{Stress-test:} The evaluation of trained models and neural architectures in a controlled way with regard to a particular type of reasoning (e.g. logic inference \cite{Richardson2019}) or linguistic capability (e.g. lexical semantics \cite{Naik2018}) is referred to as ``stress-testing'' \cite{Naik2018}. Measuring the prediction performance of a model with a particular architecture that was trained on a particular dataset on an evaluation-only stress-test \cite{Glockner2018} allows to draw conclusions about the capabilities the  model obtains from the training data. Stress-tests with a training set allow for more general conclusions whether a model with a specific architecture is capable of obtaining the capability, even when optimised with sufficient examples \cite{Kaushik2019a,Geiger2019}.

\paragraph{Robustness:} In line with the literature \cite{Wang2018,Jia2019}, we call a model ``robust'' against a method that alters the underlying (unknown) distribution of the evaluation data when compared to the training data, such as introduced by adversarial evaluation or stress-tests, if the out-of-distribution performance of the model is similar to that on the original evaluation set. The opposite of robustness is referred to as ``brittleness''.
    %\item what else is to come \ldots perhaps ``adversarial'' and ``robustness''
%\end{itemize}

\subsection{Objectives and Scope}
We aim to provide a comprehensive overview of issues in NLI data and models that are trained and evaluated upon them as well as the methodology used to report them. We set out to address the following questions:
\begin{itemize}
    \item Which NLI tasks and corresponding datasets have been investigated?
    \item Which types of weaknesses have been reported in NLI models and their training and evaluation data?
    \item What types of methods have been proposed to detect those weaknesses and their impacts on model performance and what methods have been proposed to overcome them?
    \item How have the proposed methods impacted the creation of novel datasets (that were described in published papers)?
\end{itemize}
%Further we take interest in any methods that were proposed in order to alleviate those issues.
%To this end, we collect a body of work and categorise the reported weaknesses and methods used to reveal them, with respect to their methodology, the target weakness they reveal and the dataset they are applied on. %, and their scaling and automation potential. 
%Finally, we present and discuss a synthesised summary of the findings, applications and open research questions.

%To set the scope of the survey, we limit the investigation to include methods that concern English NLI data as per the definition introduced above.

%The objectives of the survey is on the one hand to provide an overview of the methods and trends that critically investigate trainig and. On the other hand, it is desirable for a survey to identify the main research trends and gaps, open questions and propose possible future research directions.
%% Thus, we provide a comprehensive overview of methods that reveal biases in NLI data, confirm brittleness of NLI models towards specific comprehension and inference capabilities, and methods to obtain robustness towards them. 

%-- mmaaaaaaaaybe resulting from that analysis guidelines on what to pay attention to when creating a NLI dataset (``lessons learned'') 

%-- mapping of their applications to existing datasets: what has been done and what is missing 

%-- high-level summary of findings

%-- provide researchers with a overview of where the field is going and where are the gaps

\subsection{Data collection methodology}
\label{sec:collection-methodology}
To answer the first three questions we collect a literature body using the ``snowballing'' technique. Specifically, we initialise the set of surveyed papers with \newcite{gururangan2018annotation}, \newcite{Poliak2018} and \newcite{Jia2017}, because their impact helped to motivate further studies and shape the research field. For each paper in the set we follow its citations and works that have cited it according to Google Scholar and include papers that describe methods and/or their applications to report either 
(1) qualitative evaluation of training and/or test data; 
(2) superficial cues present in data and the tendency of models to pick them up;
(3) systematic issues with task formulations and/or data collection methods;
(4) analysis of specific linguistic and reasoning phenomena in data and/or models' performance on them; or
(5) enhancements of models' architecture or training procedure in order to overcome data-specific or model-specific issues, related to phenomena and cues described above.
We exclude a paper if its target task does not fall under the NLI definition established above, was published before the year 2014 or the language of the target dataset is not English; otherwise, we add it to the set of surveyed papers. With this approach, we obtain a total of 69 papers from the years 2014-2017 (6), 2018 (17), 2019 (38) and 2020 (8). More than two thirds (48) of the papers were published in venues hosted by the the Association for Computational Linguistics, whereas five and three were presented in AAAI and ICLR conferences, respectively. The remaining papers were published in other venues (3) or are available as an arXiv preprint (10).  
The papers were examined by the first author; for each paper the target task and dataset(s), the method applied and the result of the application was extracted and categorised.

To answer the final question, we took those publications introducing any of the datasets that were mentioned by at least one paper in the pool of surveyed papers and extended that collection by additional state-of-the-art NLI dataset resource papers (for detailed inclusion and exclusion criteria, see Appendix~\ref{app:dataset-corpus}). This approach yielded 73 papers. For those papers, we examine whether any of the previously collected methods were applied to report spurious correlations or whether the dataset was adversarially pruned against some model.

Although related, we deliberately do not include work that introduces adversarial attacks on NLP systems or discuss their fairness. For an overview thereof, we refer the interested reader to respective surveys conducted by \newcite{Zhang2019AdversarialSurvey} or \newcite{Xu2019AdversarialReview} for the first, and by \newcite{Mehrabi2019ALearning} for the latter.

%-- Characterise 
% 2014 = 1
% 2016 = 1
% 2017 = 4
% 2018 = 17
% 2019 = 38
% 2020 = 8

% TACL 1
% TAL 1
% *SEM 2
% SEMEVAL 1 
% COIN (@EMNLP) 1
% Ethics (@ACL) 1
% AAAI 5
% ACL 15 (8+1+4+1+1
% MRQA (@EMNLP) 3
% DeepLo @ 2019 1
% COLING 2
% CONLL 2
% EMNLP 10 (7+2+1)
% NAACL 7 (4+1+1+1)
% BlackboxNLP @ 2
% LREC 1
% ICLR 3
% arxiv 10
% ICSC 1

% ICLR: 3
% AAAI: 5
% ArXiv: 10
% ACL: 48
% others: 3

%resulting in an unavoidably incomplete survey that covers the main methods, developing trends and approaches to improve evaluation data, increase rigour in NLI evaluation and  more fine-grained understanding of models' strengths and weaknesses. 

%\subsection{Related Surveys}
%-- the one on adversarial attacks \cite{Zhang2019AdversarialSurvey}

%-- the one on bias and fairness in NLP \cite{Mehrabi2019ALearning}

%\section{Goals}
%i'm not quite sure, what i meant here
%-- find superficial clues in datasets that might be abused to reach high performance

%-- train more ``robust'' models

%-- Better ``out of distribution'' performance

%-- Better Generalisation

\section{Weaknesses in NLI data and models}
Here, we report the types of weaknesses found in state-of-the-art NLI data and models. %We give a high-level overview; for a detailed breakdown of weaknesses by dataset, we refer the reader to Appendix~\ref. 
\subsection{Data}
We identify three main types of weakness found in the data that was utilised in training and evaluating models and outline them below:
\paragraph{Spurious Correlations} 
In span extraction tasks such as MRC, question \cite{Rychalska2018}, passage wording and the position of the answer span in the passage is indicative of the expected answer for various datasets \cite{Kaushik2019}.
In the ROC stories dataset, \cite{Mostafazadeh2016} where the task is to choose the most plausible ending to a story, the endings exhibit exploitable cues \cite{Schwartz2017}.
These cues are even noticeable by humans \cite{Cai2017}.

For sentence pair classification tasks, such as RTE, \newcite{Poliak2018} and \newcite{gururangan2018annotation} showed that certain \emph{n}-grams, lexical and grammatical constructs in the hypothesis and its length correlate with the expected label for a multitude of RTE datasets. The latter study referred to these correlations as ``annotation artifacts''. \newcite{mccoy2019right} showed that lexical features like word overlap and common subsequences between the hypothesis and premise, are highly predictive of the entailment label in the MNLI dataset. Beyond RTE, the choices in the COPA \cite{roemmele2011choice} dataset, where the task is to finish a given passage (similar to ROC Stories), and ARCT \cite{Habernal2018TheWarrants} where the task is to select whether a statement warrants a claim, contain words that correlate with the expected prediction \cite{Kavumba2019,Niven2019}.

%-- syntax \cite{mccoy2019right} lexical semantics and grammaticality cues \cite{Poliak2018} 
\paragraph{Task unsuitability} \newcite{Chen2019a} demonstrated that selecting from answers in a multiple choice setting considerably simplifies the task when compared to selecting a span from the context. They further showed that for large parts of the popular \textsc{HotPotQA} dataset the answer can be found when deliberately not integrating information from multiple sentences (``multi-hop'' reasoning), replicated by \newcite{Min2019}.

\paragraph{Data Quality issues}

\newcite{Pavlick2019} argue that when training data are annotated using crowdsourcing, a fixed label representing the ground truth, usually obtained by majority vote between annotators, is not representative of the uncertainty which can be important to indicate the complexity of an example or the fact that its correctness is debateable.
Neural networks are, in fact, unable to pick up such uncertainty. 
Furthermore, both \newcite{Schlegel2020} and \newcite{Pugaliya2019} report the existence of factual errors in MRC evaluation data, where the expected answer to a question is actually wrong. Finally, \newcite{Rudinger2017} show the presence of gender and racial stereotypes in crowd-sourced RTE datasets. 
%-- factually questionable data \cite{Schlegel2020AStandards,Pugaliya2019}

\subsection{Models}
These data weaknesses contribute to brittleness in trained models themselves. Below, we outline those and other issues reported in the literature:
%which can be observed in the trained models themselves, as described below...
\paragraph{Exploitation of Cues} Given the existence of spurious correlations in NLI data, it is worthwhile knowing whether models optimised on data containing those correlations actually exploit them. 
In fact, multiple studies confirm this hypothesis, demonstrating that evaluating models on a version of the same dataset where the correlations do not exist, results in poor prediction performance \cite{mccoy2019right,Niven2019,Kavumba2019}.
\paragraph{Semantic Over-stability} Another weakness, particularly shown for MRC models, is that they appear to not capture the semantics of text beyond superficial lexical features. Neural models struggle to distinguish important from irrelevant sentences that share words with the question \cite{Jia2017}, disregard syntactic structure \cite{Basaj2018,Rychalska2018} and semantically important words \cite{Mudrakarta2018}. For RTE, they may disregard the composition of the sentence pairs \cite{Nie2019a}.
\paragraph{Generalisation Issues} Some issues hint at limited generalisation capabilities of models beyond a particular dataset. A reason lies in the typical machine learning strategy whereby data used for  evaluation is drawn from the same distribution as the training data. In the case of NLP, the distribution is determined by the design of the data collection method, usually crowd-sourced annotation of a large corpus of documents in natural language, e.g. SQuAD \cite{rajpurkar2016squad}, MNLI \cite{Williams2018}. 
A related problem is that datasets contain spurious correlations that are inherent to a particular dataset rather than to the underlying task, and that optimised models learn to exploit them as discussed above. The implications are, firstly, that models overfit to a specific dataset and do not generalise well to other examples drawn from the (unknown) task-specific distribution. Secondly, they fail to acquire linguistic and reasoning capabilities that were not explicitly required in the training sets \cite{Glockner2018,Richardson2019,Yanaka2019a}. Evaluation data drawn from the same distribution as the training data is unsuitable for revealing both of those issues.

\begin{figure}[t]
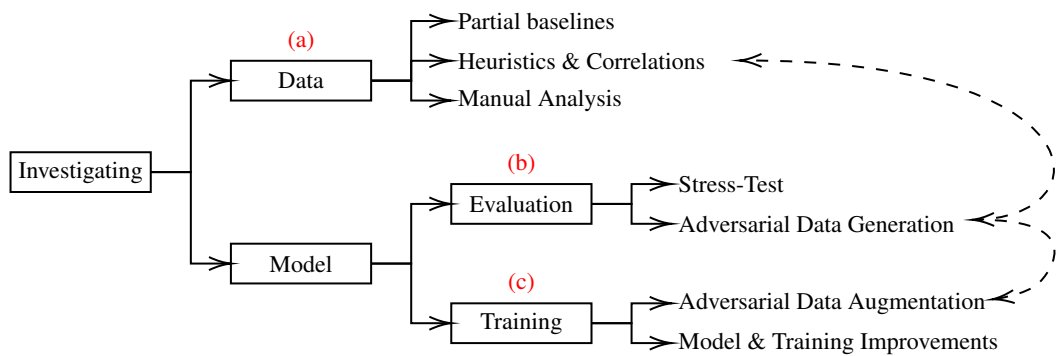

    \centering
    \include{figures/taxonomy}
    \caption{Taxonomy of investigated methods. Dashed arrows indicate conceptually related types of methods, i.e. a method of one type are commonly applied with another method of the related type. Labels (a), (b) and (c) correspond to the coarse grouping discussed in Section~\ref{sec:methods}.}
    \label{fig:taxonomy}
\end{figure}

% They can be grouped into three subcategories, namely methods that:
% \begin{enumerate}[(a)]
%     \item reveal systematic issues with existing training and evaluation data, such as the spurious correlations mentioned above
%     \item investigate what inference and reasoning capabilities models optimised on these data acquire when evaluated on samples not drawn from the training distribution, either by providing a collection of different benchmarks \cite{Dua2019} or different tasks \cite{Wang2019SuperGLUE:Systems}, or by investigating a particular capability more closely, such as processing hypo- and hypernymy \cite{Glockner2018}.
%     \item propose architectural \cite{Sagawa2019} and training procedure \cite{Wang2018} improvements in order to achieve more robust generalisation beyond data drawn from the training distribution, i.e. forcing models to ignore the biases present in training data to a certain extent.
% \end{enumerate}

\section{Methods that reveal weaknesses in NLI}
\label{sec:methods}
In the following section we categorise the surveyed papers, briefly describe the categories and illustrate the methodologies by reference to respective papers. On a high level, we distinguish between methods that (a) reveal systematic issues with existing training and evaluation data such as the spurious correlations mentioned above, (b) investigate what inference and reasoning capabilities models optimised on these data acquire when evaluated on samples not drawn from the training distribution and (c) propose architectural \cite{Sagawa2019} and training procedure \cite{Wang2018} improvements in order to achieve more robust generalisation beyond data drawn from the training distribution. A schematic overview of the taxonomy of the categories is shown in Figure~\ref{fig:taxonomy}.

%-- Classify papers according to (data-induced) taxonomy

%-- briefly describe categories and illustrate by reference to relevant papers

\subsection{Data-investigating Methods}
\label{sec:data-investigate}
Methods in this category analyse flaws in data such as cues in input that are predictive of the output \cite{gururangan2018annotation}. As training and evaluation data from state-of-the-art NLI datasets are assumed to be drawn from the same distribution, models that were fitted on those cues achieve high performance in the evaluation set, without being tested on the required inference capabilities. 
Furthermore, methods that investigate the evaluation data in order to gain a deeper understanding of the assessed capabilities \cite{Chen2016} fall under this category as well.
%It is worth noting, that when investigating training data, they are usually followed by an exhaustive evaluation of models trained on the investigated data, as detected flaws in training data do not necessitate flaws in model's performance per se. 
In the analysed body of work, we identified the following three types of methods:

\paragraph{Partial Baselines} These methods seek to verify that every input modality provided by the task is actually required to make the right prediction (e.g. both question and passage for MRC, and premise and hypothesis for RTE). Training and evaluating a classifier on parts of the input only suggests that those parts exhibit cues that correlate with the expected prediction, if the measured performance is significantly higher than randomly guessing. Both \newcite{gururangan2018annotation} and \newcite{Poliak2018} demonstrated near state-of-the-art performance on multiple RTE datasets, such as SNLI \cite{Bowman2015} and MNLI \cite{Williams2018}, when training a classifier with hypothesis-only input. \newcite{Kaushik2019} even surpass state-of-the-art MRC models on various datasets when training and evaluating only on parts of the provided input. Methods that mask, drop or shuffle input words or sentences fall under this category as well. 
Using them, \newcite{Sugawara2019} reach  performance comparable to that of a model that is trained on full input on a variety of MRC datasets. Similarly, \newcite{Nie2019a} reach near state-of-the-art performance on the SNLI and MNLI datasets when shuffling the words in the premise and hypothesis. 

Finally, we include methods here that seek to verify whether the data or task formulation is fit to evaluate a particular capability, as they involve training models that are architecturally restricted to obtain said capability, e.g. models that process documents strictly independently to answer questions that require information synthesis from multiple documents \cite{Min2019,Chen2019a}. Good performance of those impaired models indicates that the task can be solved without the required capability to a certain extent.

Above-chance performance of partial input baselines hints at spurious correlations in the data and suggests that models learn to exploit them; it does not however reveal their precise nature. The opposite does not hold true either: near-chance performance on partial input does not warrant cue-free data, as \newcite{Feng2019} illustrate on synthetic examples and published datasets.

% -- no prior knowledge required, cheap to employ

% -- input masking, substitution and perturbations also belongs to this category (e.g randomise word order, substitute for POS-tags, etc)

\paragraph{Heuristics and Correlations} These aim to unveil specific cues and spurious correlations between input and expected output that enable models to learn the task more easily. For sentence pair classification tasks, \newcite{gururangan2018annotation} use the PMI measure between words in a hypothesis and the expected label, while \newcite{Poliak2018} use the conditional probability of a label given a word. In contrast, \newcite{Tan2019} use word bigrams instead of single words to model their correlation. \newcite{mccoy2019right} count instances of (subsequently) overlapping words and mutual subtrees of the syntactic parses in a given premise and hypothesis pair, and show that their label distribution is heavily skewed towards entailment. \newcite{Nie2019a} optimise a logistic regression model on lexical features and use its confidence to predict a wrong label for a given premise-hypothesis pair as a score for the requirement of inference beyond lexical matching. \newcite{Niven2019} define \emph{productivity} and \emph{coverage} to measure how likely and for what proportion of the dataset an \emph{n}-gram is indicative of the expected label. \newcite{Cai2017} propose simple rules based on length, negation and off-the-shelf sentiment analyser scores to select the most probable ending for the ROC story completion task.

To show that models actually learn to react to the cues, the data analysis is usually followed by an evaluation on a balanced evaluation set where those correlations are not present anymore (e.g. by balancing the label distribution for a correlating cue, as described in Section~\ref{sec:eval-model}).

%-- typically word and n-gram appearance and PMI, correlation with labels (for NLI)
% \paragraph{Task Formulation Critique}     
%-- verify whether evaluation data is fit to evaluate a particular modality (e.g. information synthesis, ``Multihop'')

%-- observe performance of model that is constrained to not be able to do ``X'' on data that claims to be only solvable with ``X'' \cite{chen2019understanding,Min2019} 

%Only Question / Only Passage / Only Hypothesis / Random words / Only first sentence \cite{Jiang2019AvoidingQA,Kaushik2019,Poliak2018}

%Heuristics
%\begin{itemize}
%    \item measuring ``cues'': applicability/productivity/coverage \cite{Niven2019ProbingArguments}, insensitivity/polarity/word pairs \cite{Sanchez2018BehaviorRobustness}
%    \item Correlations: majority vote, cues etc \cite{mccoy2019right,gururangan2018annotation,Niven2019ProbingArguments,Kavumba2019}, also \cite{Poliak2018}
%\end{itemize}
\paragraph{Manual Analyses} These methods intend to qualitatively analyse the data, if automated approaches as those mentioned above are unsuitable due to the complexity of the phenomena of interest. To some extent, all papers describing experiment results on evaluation data or introducing new datasets are expected to perform a qualitative error or data analysis. We highlight a comparative qualitative analysis of state-of-the-art models on multiple MRC datasets \cite{Pugaliya2019}. Furthermore, \newcite{Schlegel2020} perform a qualitative analysis of popular MRC datasets reporting evaluated linguistic phenomena and reasoning capabilities as well as existing factual errors in data. 

%-- pick up more intricate dimensions where automatic approaches as the ones mentioned above are unsuited

%-- Error analysis (to some extent all papers that introduce a new model), more targeted: \cite{Pugaliya2019}

%-- Data analysis \cite{Yatskar2019}, also our paper

\subsection{Model-investigating Methods}
\label{sec:eval-model}
Rather than analysing data, approaches described in this section directly evaluate the models with respect to their inference capabilities with regard to various phenomena of interest. Furthermore, methods that improve a model's generalisation beyond potential biases it encounters during training, either by augmenting the training data, or by altering the architecture or the training procedure, are described here as well.

\paragraph{Stress-test} is an increasingly popular way to assess trained models and architectures. \newcite{Naik2018} automatically generate NLI evaluation data based on an analysis of observed state-of-the-art model error patterns, introducing the term ``stress-test''. Stress-tests have since been proposed to evaluate the capabilities of handling monotonicity \cite{Yanaka2019a}, lexical inference \cite{Glockner2018}, definitions \cite{Richardson2019} and compositionality \cite{Nie2019a} for RTE models and semantic equivalence \cite{Ribeiro2019} for MRC. \newcite{Liu2019}
propose an evaluation methodology to rightfully attribute the stress test performance to either missing examples in training data or the model's inherent incapability to capture the tested phenomenon by optimising the trained model on portions of the stress test data.

%-- (often automated) and controlled generation of an evaluation set focusing on a particular phenomenon (e.g. linguistic capabilities, reasoning capabilities)

%-- used to assess capabilities of trained models \cite{Yanaka2019CanReasoning,Richardson2019ProbingFragments}

%-- term ``stress-test'' coined by \cite{Naik2018StressInference}
%something something about out-of-distribution performance, if certain that phaenomenon is not present in training data

\paragraph{Adversarial Evaluation} refers to generating data with the aim to ``fool'' a target model. \newcite{Jia2017} showed that models across the leaderboard exhibit over-stability to keywords shared between a given question and passage pair in the \textsc{SQuAD} \cite{rajpurkar2016squad} dataset. These models change their prediction after the addition of distracting sentences, even if they do not alter the semantics of the passage (therefore keeping the validity of the expected answer). \newcite{Wallace2019} further showed that adversaries generated against a target model tend to be universal for a whole range of neural architectures.

Methods that evaluate whether models that are trained on data exhibiting spurious correlations inherit those, belong to this category as well.
\newcite{mccoy2019right} use patterns to generate an adversarial evaluation set with controlled distribution, such that lexical cues in the training data are not indicative of the label anymore. \newcite{Niven2019} and \newcite{Kavumba2019} add mirrored instances (i.e. modify the semantics of the sentences in a way such that the opposite label is true) of the biased data to create a set with balanced distribution of examples that contain words that otherwise correlate with the expected label in the original data.

%-- something about semantic overstability

%-- generate adversarial data to prune data-set against a certain model (e.g. data is only accepted if model fails on it)

%-- tend to be adversarial against a whole class of models, ``universal'' \cite{Jia2017,Wallace2019}

%-- used when generating data-sets \cite{Nie2019AdversarialUnderstanding,dua2019drop}
\subsection{Model-improving Methods}
Here we discuss methods that improve the robustness of models against adversarial and out-of-distribution evaluation, by either modifying the available training data or making adjustments to the training procedure.
%-- learn from biased data

%-- data-debiasing methods should go somewhere
\paragraph{Training data augmentation} methods improve the training data to train a model that is robust against a given adversary type. Thus they are inherently linked with the adversarial data generation methods. However, simply training the model on parts of the adversarial evaluation set is not always sufficient, as adversarially robust generalisation increases the sample complexity, and therefore ``requires more (training) data'' \cite{Schmidt2018AdversariallyData}. \newcite{Wang2018} introduce various improvements to the original \textsc{AddSent} algorithm, in order to generate enough training data to obtain robustness for the adversarial evaluation set introduced by \newcite{Jia2017}. \newcite{Geiger2019} propose a method to estimate the required size of the training set for any given adversarial evaluation set and apply their theory on evaluating the capability of neural networks to learn compositionality.
As an alternative to augmenting training data, \newcite{Sakaguchi2019WinoGrande:Scale} introduce \textsc{AFLite}, a method to automatically detect and remove data points that contribute to arbitrary spurious correlations. It has been since empirically validated and theoretically underpinned by \newcite{Bras2020}.

Furthermore, we include the application of adversarial data generation when employed during the construction of a new dataset: in crowd-sourcing, where humans act as adversary generators and an entry is only accepted if it triggers a wrong prediction by a trained target model \cite{Nie2019,Dua2019}, or when automatically generating multiple choice alternatives until a target model cannot distinguish between human-written and automatically generated options, called \emph{Adversarial Filtering} \cite{Zellers2018,Zellers2019}.

%-- linked to evaluation on adversarialt data (obviously)

\paragraph{Architecture and Training Procedure Improvements} deviate from the idea of data augmentation and seek to train robust and de-biased models from potentially biased data. These methods include joint training (and discarding) of robust models together with models that are designed to exploit the dataset biases \cite{Clark2019,He2019}, re-weighting the loss function to incorporate the bias in the data \cite{Schuster2019,Zhang2019}, parameter regularisation \cite{Sagawa2019} and the use of external resources, such as linguistic knowledge \cite{Zhou2019,Wu2019} or logic \cite{Minervini2018}.

%-- rather than augmenting data, changing training procedure or model architecture to obtain robustness against an adversarial dataset

%-- e.g. train ensemble of models \cite{Clark2019}, re-balance batch loss \cite{Zhang2019MitigatingAbility}, whatever \cite{Sagawa2019DistributionallyGeneralization} and \cite{He2019UnlearnResidual} do \ldots 

%\cite{Nie2019AdversarialUnderstanding,dua2019drop,Glockner2018BreakingInferences}
%    Evaluation: generate adversaries
%            item automatically: a lot of adversarial attacks, particularly \cite{Wallace2019} for NLP
%            item With expert knowledge: \cite{mccoy2019right,Jia2017,wang2018robust,Jiang2019AvoidingQA}

%Social biases
%\begin{itemize}
%    \item mostly gender bias in Coreference Resolution datasets \cite{Zhao2018GenderMethods,Rudinger2018GenderResolution,Webster2018}
%    \item \ldots and NLI \cite{Rudinger2017SocialInferences}
%\end{itemize}
%jointly optimise ``smart'' model alongside with ``biased'' model i.e. one that is incentivised to learn bias data, perform inference with ``smart'' model only \cite{Clark2019DontBiases} 

%\section{Tasks}
%-- Nli

%-- QA/MRC

%-- VQA? not sure if we want to tap into that, but there's a lot of work there
%\section{Expert Knowledge and Automation potential}
%Expertise:
%\begin{itemize}
%    \item Machine Learning
%    \item Linguistics
%    \item ...
%    \item None
%\end{itemize}

%Automation:
%\begin{itemize}
%    \item Full
%    \item HIL
%    \item Semi-automated
%    \item None
%\end{itemize}
%-- full

%-- none (manual)
\section{Results and Discussion}
%\begin{figure}
%    \centering
%    \include{figures/by_task}
%    \caption{Distribution of tasks corresponding to analysed datasets or models}
%    \label{fig:by-task}
%\end{figure}

% \begin{table}[b]
% \centering
% \begin{tabularx}{1\columnwidth}{Y X}
% \hline
% Name/Source & Target Phenomenon\\
% \hline
% \multicolumn{2}{l}{\textbf{RTE}} \\
% \hline
% HANS \newline \cite{mccoy2019right} & data cleaned from spurious correlations observed in the MNLI dataset \\
% \hline
% \multicolumn{2}{l}{\textbf{MRC}} \\
% \hline
% SQuAD-Adv \newline \cite{Jia2017} & Questions and passages generated adversarially for models trained on the SQuAD dataset \\ 
% \hline
% \multicolumn{2}{l}{\textsc{Other}} \\
% \hline
% \end{tabularx}
% \caption{Resources for fine-grained model evaluation .}
% \label{table:selected-datasets}
% \end{table}

\begin{figure}[t]
    \centering
    \begin{minipage}[t]{0.54\textwidth}
        %\captionsetup{width=.9\linewidth}
        \include{figures/by_category}
        \vspace{-2.1\baselineskip}
        \caption{Number of methods per category split by task. As multiple papers report more than one method, the maximum (86) does not add up to the number of surveyed papers (69).}
        \label{fig:by-category}
    \end{minipage}
    \quad
    \begin{minipage}[t]{0.43\textwidth}
    %\captionsetup{width=.9\linewidth}
    \centering
    \include{figures/by_year}
    \vspace{-1.5\baselineskip}
    \caption{Dataset by publication year with {\color{red}no} or {\color{blue}any} spurious correlations detection methods applied; applied in a  {\color{orange}later} publication; created using {\color{green}adversarial} filtering, or {\color{yellow}both}.}
    \label{fig:by-year}
    \end{minipage}
\end{figure}

%FIGURES
% \begin{figure}[t]
%     \centering
% \begin{subfigure}[t]{0.54\textwidth}
%     \centering
%     \captionsetup{width=.9\linewidth}
%     \include{figures/by_category}
%     \vspace{-2\baselineskip}
%     \caption{Number of methods per category split by task. As multiple papers report more than one method, the maximum (86) does not add up to the number of surveyed papers (69)}
%     \label{fig:by-category}
% \end{subfigure}%    
% \begin{subfigure}[t]{0.46\textwidth}
%     \captionsetup{width=.9\linewidth}
%     \centering
%     \include{figures/by_year}
%     \vspace{-1.4\baselineskip}
%     \caption{Dataset by year of publication with {\color{red}no} or {\color{blue}any} spurious correlations detection methods applied, those where they were applied in a  {\color{orange}later} publication, those created using {\color{green}adversarial} filtering, or {\color{yellow}both}.}
%     \label{fig:by-year}
% \end{subfigure}
% %\begin{subfigure}[t]{1\textwidth}
% %    \centering
% %    \includegraphics[width=1\textwidth]{figures/datasets.png}
% %    \caption{Word cloud with investigated datasets. Size proportional to the number of surveyed papers %investigating the dataset.}
% %    \label{fig:by-dataset}
% %\end{subfigure}
% \caption{Visualisation of the survey results.}
% \label{fig:results}
% \end{figure}

We report the result of our categorisation of the literature in this section. %As illustrated in Figure~\ref{fig:by-task}, 
More than half of the surveyed papers (35) are focusing on the RTE task, followed by analysis of the MRC (25) task with 4 and 5 investigating other and multiple tasks, respectively. Looking at the breakdown by type of analysis according to our taxonomy (Figure~\ref{fig:by-category}) we see that most approaches concern adversarial evaluation and propose improvements for robustness against biased data and adversarially generated test data. This is not surprising, as robustness against a type of adversary can only be empirically validated via evaluation on the corresponding adversarial test set.

It is worth highlighting that there is little work analysing MRC data with regard to spurious correlations. We attribute this to the fact, that it is hard to conceptualise the correlations of input and expected output for MRC beyond very coarse heuristics (such as sentence position or lexical answer type), as the input is a whole paragraph and a question and the expected output is typically a span anywhere in the paragraph. For RTE, by contrast, where the input consists of two sentences and the expected output is one of three fixed class labels, possible correlations are easier to unveil.
In fact, the sole paper (included in our survey) which reports spurious correlations in MRC data, investigated a dataset where the goal is to predict the right answer given four alternatives, thus considerably constraining the expected output space \cite{Yu2020}. Finally, there are few (4) “stress-tests” for the task of MRC. Those focus on prediction consistency \cite{Ribeiro2019}, acquired knowledge \cite{Richardson2019}, unanswerability  \cite{Nakanishi2018} or multi-dataset evaluation \cite{Dua2019a} rather than performing an analysis of acquired linguistic or reasoning capabilities.

%-- plot: by target task

%-- plot: by taxonomy

%-- plot: by target dataset

%Figure~\ref{fig:by-dataset} illustrates the datasets and models trained on those datasets that were analysed in the surveyed papers.
Regarding the datasets used in the surveyed papers most analyses were done on the SNLI and MNLI datasets (20 and 22 papers, respectively) For RTE. For MRC, the most analysed dataset is \textsc{SQuAD}. %Figure~\ref{fig:by-dataset} illustrates the datasets and models trained on those datasets that were analysed in the surveyed papers. For RTE, most analyses were done on the SNLI and MNLI datasets (20 and 22 papers, respectively). For MRC, the most analysed dataset is \textsc{SQuAD} (17). 
17 RTE and 30 MRC datasets were analysed at least once; we attribute the difference to the existence of various different MRC datasets and the tendency of performing multi-dataset analyses in papers that investigate MRC datasets \cite{Kaushik2019,Sugawara2019,Si2019}. For a full list of investigated datasets and the weaknesses reported on them, please refer to Appendix~\ref{app:detailed-results}.

%- plot with new datasets vs how many of them had any of the reviewed methods applied to analyse them when dataset was introduced

%-- something about the target datasets. RTE dominates with SNLI and MNLI, many MRC datasets, because of multiple MRC benchmarking suites

%-- observation: more RTE than anything else (particularly MRC)

%-- observation: (``few'') target datasets

%To answer the final question %of how the proposed methods influence the creation of novel datasets, we took those publications introducing any of the datasets that were mentioned by at least one paper in the pool of surveyed papers and extended that collection by additional state-of-the-art NLI dataset resource papers (for detailed inclusion and exclusion criteria and the coding schema, see Appendix~??).
We report, whether the existence of spurious correlations was investigated in the original or a later publication, by applying quantitative methods such as those discussed in Section~\ref{sec:data-investigate}: Partial Baselines and Heuristics and Correlations, or whether the dataset was generated adversarially against a neural model. The results are shown in Figure \ref{fig:by-year}. We observe that the publications we use as our seed papers for the survey (c.f. Section~\ref{sec:collection-methodology}) in fact seem to impact how novel datasets are presented, as after their publication (in years 2017 and 2018) a growing number of papers report partial baseline results and advanced correlations in their data (three in 2018 and seven in 2019). Furthermore, newly proposed resources are progressively pruned against neural models (eight in 2018 and 2019 cumulative). However, for nearly a half (36 out of 75) of the datasets under investigation there is no information about potential spurious correlations and biases yet.

A noteworthy corollary of the survey is that -- perhaps unsurprisingly -- neural models' notion of complexity does not necessarily correlate with that of humans. In fact, after creating a ``hard'' subset of their evaluation data that is clean of correlations, \newcite{Yu2020} report a better human performance than on the biased version, directly contrary to neural models they evaluate. Partial baseline methods suggest a similar conclusion: without the help of statistics, humans will arguably not be able to infer, whether a sentence is entailed by another sentence they never see, whereas neural networks excel at it \cite{Poliak2018,gururangan2018annotation}.
Additionally, models' prediction confidence does not correlate with human confidence as approximated by inter-annotator agreement on a variety of RTE datasets \cite{Pavlick2019}.

Finally, results suggest that models can benefit from different types of knowledge that enables them to learn to perform the task even when trained on biased data. Models that incorporate structural biases \cite{Battaglia2018}, e.g. by operating on syntax trees rather than plain text, are more robust to syntactic adversaries \cite{mccoy2019right}. In the case of models that build upon large pre-trained language models, the number of the parameters and the size of the corpus used for language model training appear beneficial \cite{Kavumba2019}.

%-- open RQ: correlation of pre-training corpus size + model size to bad performance on "out of distribution" stress-tests/adversarial tests \cite{Kavumba2019} hints that larger models (e.g. roberta) perform better on out-of-distribution stress-test, but again, no empirical results as of yet

%-- open RQ: correlation of inductive biases (cite what is inductive bias) to model performance on "out of distribution" tests \cite{mccoy2019right} results hint, but no large-scale empirical study conducted yet, to our knowledge.

\section{Conclusion}
We present a structured survey of methods that reveal heuristics and spurious correlations in datasets, methods which show that neural models inherit those correlations or assess their capabilities otherwise, and methods that mitigate this by adversarial training, data augmentation and model architecture or training procedure improvements. 
Various NLI datasets are reported to contain spurious correlations between input and expected output,  might be unsuitable to evaluate some task modality due to dataset design or suffer from quality issues. RTE is a popular target task for these data-centred investigations with more than half of the surveyed papers focusing on it.
NLI models, in turn, are shown to exploit those correlations and to rely on superficial lexical cues. Furthermore, they lack generalisation beyond the evaluation set resulting in poor performance on out-of-distribution evaluation sets, generated adversarially or targeted at a specific capability. Efforts to achieve robustness include augmenting the training data with adversarial examples, making use of external resources and modifying the neural network architecture or training objective.
%-- data weaknesses: spurious correlations, task unsuitability, data quality issues; -- model weaknesses: cue reliance over-stability and weak ood generalisation -- RTE datasets popular target for those investigations. 
%about half of recently proposed datasets are still not investigated
% We note that -- RTE task mostly analysed for cues and correlations -- used mostly for stress-testing models for various capabilities 
%there is little work done on automatically detecting heuristics and cues in MRC training data. A possible explanation for this is the fact that RTE data is easier to analyse, as the output space is more constrained compared to MRC. Finally, there are but few ``stress-tests'' for the task of MRC, particularly with regard to linguistic capabilities. 

Based on these findings, we formulate the following recommendations for possible future research directions:
\begin{itemize}
    \item There is a need for an empirical study that systematically investigates the benefits of type and amount of prior knowledge on neural models' out-of-distribution stress test performance. 

    \item We believe the scientific community will benefit from an application of the quantitative methods that have been presented in this survey to the remaining 36 recently proposed NLI datasets that have not been examined for spurious correlations yet.
    
    \item Partial baselines are conceptually simple and cheap to employ for any given task, so we want to incentivise researchers to apply and report their performance, when introducing a novel dataset. While not a guarantee for the absence of spurious correlations \cite{Feng2019}, they can hint at their presence and serve as an upper bound for the complexity of the dataset.

    \item  Adapting methods applied to RTE datasets or developing novel methodology to reveal cues and spurious correlations in MRC data is a possible future research direction.

    \item While RTE is increasingly becoming a proxy task to attribute various reading and reasoning capabilities to neural models, the transfer of those capabilities to different tasks, such as MRC, remains to be shown yet. Additionally, the MRC task requires further capabilities that cannot be tested in an RTE setting conceptually, such as selecting the relevant answer sentence from distracting context or integrating information from multiple sentences, both shown to be inadequately tested by current state-of-the-art gold standards \cite{Jia2017,Jiang2019}. Therefore it is important to develop those ``stress-tests'' for  MRC models as well, in order to gain a more focussed understanding of their capabilities and limitations.

\end{itemize}

We want to highlight, that albeit exhibiting cues or weaknesses in design, the availability of multiple large-scale datasets is a vital step in order to gain empirically grounded understanding of what the current state-of-the-art NLI models are learning and where they still fail. This is a necessary requirement for building the next iteration of datasets and model architectures and therefore further advance the reseach in NLP. %, blah blah blah, we don't want to criticise, blah blah blah (maybe even in introduction) 

While the discussed methods seem to be necessary to make progress and gain a precise understanding of the capabilities and, most importantly, of the limits of existing (deep learning-based) approaches and can guide research towards solving the NLI task beyond leaderboard performance on a single dataset, the question persists whether they are sufficient. It remains to be seen whether the availability of benchmark suites \cite{Wang2019,Wang2019b} consisting of multiple training and evaluation datasets -- open-domain or targeted at a specific phenomenon -- will provide enough diversity to optimise models that are robust enough to perform any given natural language understanding task, the so called ``general linguistic intelligence'' \cite{Yogatama2019LearningIntelligence}.

%-- Same for robustness, we can only certify against adversarial methods we know of

%-- or whether we need to research new model classes sth sth hybrid, sth sth causal inference etc

%-- yeah, these are the RQ that generally drive AI research, so i don't know if we should talk about it here

%-- oh, also: 
\bibliographystyle{acl}
\bibliography{references}
\newpage
\appendix
\include{appendix/appendix-datasets}
\newpage
\include{appendix/inclusion-criteria}
\end{document}

%% file: figures/taxonomy.tex
\tikzset{every picture/.style={line width=0.75pt}} %set default line width to 0.75pt        

\begin{tikzpicture}[x=0.75pt,y=0.75pt,yscale=-1,xscale=1]
%uncomment if require: \path (0,181.39999389648438); %set diagram left start at 0, and has height of 181.39999389648438

%Shape: Rectangle [id:dp7980951273108542] 
\draw   (0,77) -- (70,77) -- (70,97) -- (0,97) -- cycle ;
%Shape: Rectangle [id:dp4363064241982936] 
\draw   (110,31) -- (180,31) -- (180,51) -- (110,51) -- cycle ;
%Shape: Rectangle [id:dp004663179707912302] 
\draw   (110,123) -- (180,123) -- (180,143) -- (110,143) -- cycle ;
%Shape: Rectangle [id:dp7219122195756015] 
\draw   (220,93) -- (290,93) -- (290,113) -- (220,113) -- cycle ;
%Shape: Rectangle [id:dp783287532775125] 
\draw   (220,153) -- (290,153) -- (290,173) -- (220,173) -- cycle ;
%Straight Lines [id:da027976835213297635] 
\draw    (70,87) -- (90,87) -- (90,41) -- (108,41) ;
\draw [shift={(110,41)}, rotate = 180] [color={rgb, 255:red, 0; green, 0; blue, 0 }  ][line width=0.75]    (10.93,-3.29) .. controls (6.95,-1.4) and (3.31,-0.3) .. (0,0) .. controls (3.31,0.3) and (6.95,1.4) .. (10.93,3.29)   ;
%Straight Lines [id:da600152596164515] 
\draw    (70,87) -- (90,87) -- (90,133) -- (108,133) ;
\draw [shift={(110,133)}, rotate = 180] [color={rgb, 255:red, 0; green, 0; blue, 0 }  ][line width=0.75]    (10.93,-3.29) .. controls (6.95,-1.4) and (3.31,-0.3) .. (0,0) .. controls (3.31,0.3) and (6.95,1.4) .. (10.93,3.29)   ;
%Straight Lines [id:da4078314770281336] 
\draw    (180,133) -- (200,133) -- (200,103) -- (218,103) ;
\draw [shift={(220,103)}, rotate = 180] [color={rgb, 255:red, 0; green, 0; blue, 0 }  ][line width=0.75]    (10.93,-3.29) .. controls (6.95,-1.4) and (3.31,-0.3) .. (0,0) .. controls (3.31,0.3) and (6.95,1.4) .. (10.93,3.29)   ;
%Straight Lines [id:da6915865390966427] 
\draw    (180,133) -- (200,133) -- (200,163) -- (218,163) ;
\draw [shift={(220,163)}, rotate = 180] [color={rgb, 255:red, 0; green, 0; blue, 0 }  ][line width=0.75]    (10.93,-3.29) .. controls (6.95,-1.4) and (3.31,-0.3) .. (0,0) .. controls (3.31,0.3) and (6.95,1.4) .. (10.93,3.29)   ;
%Straight Lines [id:da8829561756383693] 
\draw    (180,41) -- (200,41) -- (200,11) -- (218,11) ;
\draw [shift={(220,11)}, rotate = 180] [color={rgb, 255:red, 0; green, 0; blue, 0 }  ][line width=0.75]    (10.93,-3.29) .. controls (6.95,-1.4) and (3.31,-0.3) .. (0,0) .. controls (3.31,0.3) and (6.95,1.4) .. (10.93,3.29)   ;
%Straight Lines [id:da05687819234102165] 
\draw    (180,41) -- (200,41) -- (200,31) -- (218,31) ;
\draw [shift={(220,31)}, rotate = 180] [color={rgb, 255:red, 0; green, 0; blue, 0 }  ][line width=0.75]    (10.93,-3.29) .. controls (6.95,-1.4) and (3.31,-0.3) .. (0,0) .. controls (3.31,0.3) and (6.95,1.4) .. (10.93,3.29)   ;
%Straight Lines [id:da445950670490733] 
\draw    (180,41) -- (200,41) -- (200,51) -- (218,51) ;
\draw [shift={(220,51)}, rotate = 180] [color={rgb, 255:red, 0; green, 0; blue, 0 }  ][line width=0.75]    (10.93,-3.29) .. controls (6.95,-1.4) and (3.31,-0.3) .. (0,0) .. controls (3.31,0.3) and (6.95,1.4) .. (10.93,3.29)   ;
%Straight Lines [id:da7139682857352219] 
%\draw    (180,41) -- (200,41) -- (200,71) -- (218,71) ;
%\draw [shift={(220,71)}, rotate = 180] [color={rgb, 255:red, 0; green, 0; blue, 0 }  ][line width=0.75]    (10.93,-3.29) .. controls (6.95,-1.4) and (3.31,-0.3) .. (0,0) .. controls (3.31,0.3) and (6.95,1.4) .. (10.93,3.29)   ;
%Straight Lines [id:da6376872294401805] 
\draw    (290,103) -- (310,103) -- (310,93) -- (328,93) ;
\draw [shift={(330,93)}, rotate = 180] [color={rgb, 255:red, 0; green, 0; blue, 0 }  ][line width=0.75]    (10.93,-3.29) .. controls (6.95,-1.4) and (3.31,-0.3) .. (0,0) .. controls (3.31,0.3) and (6.95,1.4) .. (10.93,3.29)   ;
%Straight Lines [id:da10801905497493225] 
\draw    (290,103) -- (310,103) -- (310,113) -- (328,113) ;
\draw [shift={(330,113)}, rotate = 180] [color={rgb, 255:red, 0; green, 0; blue, 0 }  ][line width=0.75]    (10.93,-3.29) .. controls (6.95,-1.4) and (3.31,-0.3) .. (0,0) .. controls (3.31,0.3) and (6.95,1.4) .. (10.93,3.29)   ;
%Straight Lines [id:da4159486447820546] 
\draw    (290,163) -- (310,163) -- (310,153) -- (328,153) ;
\draw [shift={(330,153)}, rotate = 180] [color={rgb, 255:red, 0; green, 0; blue, 0 }  ][line width=0.75]    (10.93,-3.29) .. controls (6.95,-1.4) and (3.31,-0.3) .. (0,0) .. controls (3.31,0.3) and (6.95,1.4) .. (10.93,3.29)   ;
%Straight Lines [id:da4701946010963137] 
\draw    (290,163) -- (310,163) -- (310,173) -- (328,173) ;
\draw [shift={(330,173)}, rotate = 180] [color={rgb, 255:red, 0; green, 0; blue, 0 }  ][line width=0.75]    (10.93,-3.29) .. controls (6.95,-1.4) and (3.31,-0.3) .. (0,0) .. controls (3.31,0.3) and (6.95,1.4) .. (10.93,3.29)   ;
%Curve Lines [id:da2884846791678287] 
\draw  [dash pattern={on 4.5pt off 4.5pt}]  (492,150.72) .. controls (500.08,149.5) and (506.46,147.31) .. (511.17,144.56) .. controls (516.2,141.61) and (542.4,110.5) .. (481.86,110.98) ;
\draw [shift={(480,111)}, rotate = 358.97] [color={rgb, 255:red, 0; green, 0; blue, 0 }  ][line width=0.75]    (10.93,-3.29) .. controls (6.95,-1.4) and (3.31,-0.3) .. (0,0) .. controls (3.31,0.3) and (6.95,1.4) .. (10.93,3.29)   ;
\draw [shift={(490,151)}, rotate = 352.77] [color={rgb, 255:red, 0; green, 0; blue, 0 }  ][line width=0.75]    (10.93,-3.29) .. controls (6.95,-1.4) and (3.31,-0.3) .. (0,0) .. controls (3.31,0.3) and (6.95,1.4) .. (10.93,3.29)   ;
%Curve Lines [id:da5571356015004257] 
\draw  [dash pattern={on 4.5pt off 4.5pt}]  (480,111) .. controls (547.93,110.87) and (544.93,29.87) .. (363.93,30.87) ;
\draw [shift={(363.93,30.87)}, rotate = 359.68] [color={rgb, 255:red, 0; green, 0; blue, 0 }  ][line width=0.75]    (10.93,-3.29) .. controls (6.95,-1.4) and (3.31,-0.3) .. (0,0) .. controls (3.31,0.3) and (6.95,1.4) .. (10.93,3.29)   ;

% Text Node
\draw (222,11) node [anchor=west] [inner sep=0.75pt]  [font=\small] [align=left] {Partial baselines};
% Text Node
\draw (222,31) node [anchor=west] [inner sep=0.75pt]  [font=\small] [align=left] {{\fontfamily{ptm}\selectfont Heuristics \& Correlations}};
% Text Node
\draw (222,51) node [anchor=west] [inner sep=0.75pt]  [font=\small] [align=left] {Manual Analysis};
% Text Node
%\draw (222,71) node [anchor=west] [inner sep=0.75pt]  [font=\small] [align=left] ;
% Text Node
\draw (332,93) node [anchor=west] [inner sep=0.75pt]  [font=\small] [align=left] {Stress-Test};
% Text Node
\draw (332,113) node [anchor=west] [inner sep=0.75pt]  [font=\small] [align=left] {{\fontfamily{ptm}\selectfont {\small Adversarial Data Generation}}};
% Text Node
\draw (332,153) node [anchor=west] [inner sep=0.75pt]  [font=\small] [align=left] {{\fontfamily{ptm}\selectfont {\small Adversarial Data Augmentation}}};
% Text Node
\draw (332,173) node [anchor=west] [inner sep=0.75pt]  [font=\small] [align=left] {Model \& Training Improvements};
% Text Node
\draw (35,87) node  [font=\small] [align=left] {{\small Investigating}};
% Text Node
\draw (145,21) node  [font=\small] [align=left] {\small {\color{red} (a)}};
\draw (145,41) node  [font=\small] [align=left] {{\small Data}};
% Text Node
\draw (145,133) node  [font=\small] [align=left] {{\small Model}};
% Text Node
\draw (255,83) node  [font=\small] [align=left] {\small {\color{red} (b)}};
\draw (255,103) node  [font=\small] [align=left] {{\small Evaluation}};
% Text Node
\draw (255,143) node  [font=\small] [align=left] {\small {\color{red} (c)}};
\draw (255,163) node  [font=\small] [align=left] {{\small Training}};

\end{tikzpicture}

%% file: figures/by_category.tex
\begin{tikzpicture}
\begin{axis}[
    ybar stacked,
    height=11em,
    width  = \textwidth,
    ymin=0,
    ymax=27,
    scaled y ticks = false,
    bar width=8pt,
    x tick label style={rotate=30, anchor=north east, align=right,text width=2cm, font=\scriptsize\itshape},
    major x tick style = transparent,
    ylabel=\emph{\# papers},
    xmajorgrids=true,
    x tick label as interval,
    ymajorgrids=true,
    y grid style=dashed,
    legend pos=north west,
    legend cell align={left},
    legend columns=2,
    xtick={0, 1, 2, 3, 4, 5, 6, 7},
    xmin=0, xmax=7,
    xticklabels={Partial Baselines, Heuristics, Manual Analyses, Stress-test, Adversarial Evaluation, Data Improvements, Arch/Training Improvements},
]
\addplot[legend entry=\textsc{MRC}, color=rred, fill=rred]  coordinates {(0.5,4) (1.5,1) (2.5,4) (3.5,4) (4.5,9) (5.5,3) (6.5,4)};
\addplot[legend entry=\textsc{Multiple}, color=pink, fill=pink]  coordinates {(0.5,1) (1.5,0) (2.5,0) (3.5,1) (4.5,1) (5.5,0) (6.5,2)};
\addplot[legend entry=\textsc{Other}, color=ggreen, fill=ggreen]  coordinates {(0.5,2) (1.5,3) (2.5,0) (3.5,1) (4.5,1) (5.5,0) (6.5,0)};
\addplot[legend entry=\textsc{RTE}, color=bblue, fill=bblue]  coordinates {(0.5,6) (1.5,9) (2.5,1) (3.5,7) (4.5,6) (5.5,6) (6.5,10)};

%\begin{pgfonlayer}{background}
%   \fill[color=black!10] (axis cs:0,0) rectangle (axis cs:4,20);
%\end{pgfonlayer}
\end{axis}
\end{tikzpicture}

%% file: figures/by_year.tex
\begin{tikzpicture}
\begin{axis}[
    ybar stacked,
    height=11em,
    width  = \textwidth,
    ymin=0,
    ymax=27,
    legend pos=north west,
    scaled y ticks = false,
    bar width=8pt,
    x tick label style={rotate=30, anchor=north east, align=right,text width=2cm, font=\scriptsize\itshape},
    major x tick style = transparent,
    ylabel={\emph{\# datasets}},
    xmajorgrids=true,
    x tick label as interval,
    ymajorgrids=true,
    y grid style=dashed,
    legend columns=2,
    xtick={0, 1, 2, 3, 4, 5, 6},
    xmin=0, xmax=6,
    xticklabels={2015, 2016, 2017, 2018, 2019, 2020},
    ]

%\legend{adv,both,late,no,yes}
    \addplot[legend entry=\textsc{no}, color=red, fill=red]  coordinates {(0.5,3) (1.5,4) (2.5,5) (3.5,10) (4.5,14) (5.5,0)};
\addplot[legend entry=\textsc{later}, color=orange, fill=orange]  coordinates {(0.5,5) (1.5,4) (2.5,3) (3.5,10) (4.5,2) (5.5,0)};
\addplot[legend entry=\textsc{yes}, color=blue, fill=blue]  coordinates {(0.5,0) (1.5,0) (2.5,0) (3.5,3) (4.5,3) (5.5,1)};
\addplot[legend entry=\textsc{both}, color=yellow, fill=yellow]  coordinates {(0.5,0) (1.5,0) (2.5,0) (3.5,0) (4.5,3) (5.5,0)};
\addplot[legend entry=\textsc{adv}, color=green, fill=green]  coordinates {(0.5,0) (1.5,0) (2.5,0) (3.5,1) (4.5,4) (5.5,0)};

\end{axis}
\end{tikzpicture}

%% file: appendix/appendix-datasets.tex
\section{Detailed Survey Results}
\label{app:detailed-results}
\begin{figure}[h]
    %\centering
    \includegraphics[width=1\textwidth]{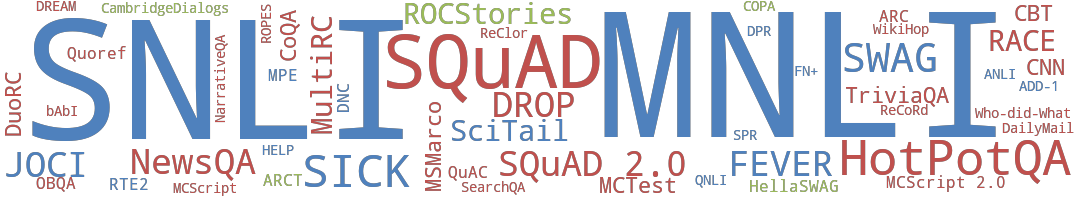}
    \caption{Word cloud with investigated {\color{bblue}RTE}, {\color{rred}MRC} and {\color{ggreen}other} datasets. Size proportional to the number of surveyed papers investigating the dataset.}
    \label{fig:by-dataset}
\end{figure}

The following table shows the full list of surveyed papers, grouped by dataset and method applied. As papers might report the application of multiple methods on multiple datasets, they can appear in the table more than once.
%\begin{table}[h]
    %\centering
    \begin{longtable}{ p{.11\textwidth} | p{.24\textwidth} | p{.57\textwidth} }
         \textbf{Dataset} & \textbf{Method used} & \textbf{Used by / Investigated by} \\
         \hline
         HotPotQA & Partial Baselines & \cite{Min2019,Sugawara2019,chen2019understanding} \\
 & Adversarial Evaluation & \cite{Jiang2019} \\
 & Data Improvements & \cite{Jiang2019} \\
 & Arch/Training Improvements & \cite{Jiang2019} \\
 & Manual Analyses & \cite{Schlegel2020,Pugaliya2019} \\
\hline
MNLI & Stress-test & \cite{Naik2018,Glockner2018,mccoy2019right,Liu2019,Nie2019a,Richardson2019a} \\
 & Arch/Training Improvements & \cite{Wang2019,He2019,Sagawa2019,Minervini2018,Mahabadi2019,Zhang2019,Clark2019,Mitra2019,Yaghoobzadeh2019} \\
 & Heuristics & \cite{gururangan2018annotation,Poliak2018,mccoy2019right,Zhang2019a,Nie2019a,Bras2020,Tan2019} \\
 & Partial Baselines & \cite{gururangan2018annotation,Poliak2018,Nie2019a} \\
 & Manual Analyses & \cite{Pavlick2019} \\
 & Adversarial Evaluation & \cite{Chien2019,Nie2019a} \\
 & Data Improvements & \cite{Mitra2019} \\
\hline
HELP & Data Improvements & \cite{Yanaka2019} \\
\hline
SNLI & Stress-test & \cite{Glockner2018,Nie2019a,Richardson2019a} \\
 & Data Improvements & \cite{Kang2018,Mitra2019,Kaushik2019a} \\
 & Heuristics & \cite{gururangan2018annotation,Poliak2018,Zhang2019a,Nie2019a,Rudinger2017,Bras2020,Tan2019} \\
 & Partial Baselines & \cite{gururangan2018annotation,Poliak2018,Feng2019,Nie2019a} \\
 & Adversarial Evaluation & \cite{Sanchez2018,Nie2019a} \\
 & Manual Analyses & \cite{Pavlick2019} \\
 & Arch/Training Improvements & \cite{He2019,Minervini2018,Mahabadi2019,Zhang2019,Jia2019,Mitra2019} \\
\hline
SciTail & Stress-test & \cite{Glockner2018} \\
 & Heuristics & \cite{Poliak2018} \\
 & Partial Baselines & \cite{Poliak2018} \\
\hline
COPA & Heuristics & \cite{Kavumba2019} \\
 & Stress-test & \cite{Kavumba2019} \\
\hline
SICK & Arch/Training Improvements & \cite{Wang2019,Zhang2019} \\
 & Heuristics & \cite{Poliak2018,Zhang2019a} \\
 & Partial Baselines & \cite{Poliak2018,Lai2015} \\
\hline
ADD-1 & Heuristics & \cite{Poliak2018} \\
 & Partial Baselines & \cite{Poliak2018} \\
\hline
DPR & Heuristics & \cite{Poliak2018} \\
 & Partial Baselines & \cite{Poliak2018} \\
\hline
FN+ & Heuristics & \cite{Poliak2018} \\
 & Partial Baselines & \cite{Poliak2018} \\
\hline
JOCI & Heuristics & \cite{Poliak2018} \\
 & Partial Baselines & \cite{Poliak2018} \\
 & Manual Analyses & \cite{Pavlick2019} \\
 & Arch/Training Improvements & \cite{Zhang2019} \\
\hline
MPE & Heuristics & \cite{Poliak2018} \\
 & Partial Baselines & \cite{Poliak2018} \\
\hline
SPR & Heuristics & \cite{Poliak2018} \\
 & Partial Baselines & \cite{Poliak2018} \\
\hline
SQuAD & Adversarial Evaluation & \cite{Rychalska2018,Wallace2019,Mudrakarta2018,Jia2017,Basaj2018} \\
 & Arch/Training Improvements & \cite{Min2018,Wu2019,Zhou2019,Clark2019} \\
 & Stress-test & \cite{Liu2019,Dua2019a,Nakanishi2018,Ribeiro2019} \\
 & Data Improvements & \cite{Wang2018,Nakanishi2018} \\
 & Partial Baselines & \cite{Sugawara2019,Kaushik2019} \\
 & Manual Analyses & \cite{Pugaliya2019} \\
\hline
DROP & Adversarial Evaluation & \cite{Dua2019} \\
 & Manual Analyses & \cite{Schlegel2020} \\
 & Stress-test & \cite{Dua2019a} \\
\hline
DNC & Manual Analyses & \cite{Pavlick2019} \\
\hline
RTE2 & Manual Analyses & \cite{Pavlick2019} \\
\hline
MSMarco & Manual Analyses & \cite{Schlegel2020,Pugaliya2019} \\
\hline
MultiRC & Manual Analyses & \cite{Schlegel2020} \\
 & Partial Baselines & \cite{Sugawara2019} \\
\hline
NewsQA & Manual Analyses & \cite{Schlegel2020} \\
 & Arch/Training Improvements & \cite{Min2018} \\
 & Stress-test & \cite{Dua2019a} \\
\hline
ReCoRd & Manual Analyses & \cite{Schlegel2020} \\
\hline
ROCStories & Partial Baselines & \cite{Schwartz2017,Cai2017} \\
 & Heuristics & \cite{Cai2017} \\
\hline
TriviaQA & Arch/Training Improvements & \cite{Min2018,Clark2019} \\
\hline
FEVER & Arch/Training Improvements & \cite{Mahabadi2019,Schuster2019} \\
 & Adversarial Evaluation & \cite{Thorne2019} \\
 & Heuristics & \cite{Schuster2019} \\
 & Data Improvements & \cite{Schuster2019} \\
\hline
ARCT & Heuristics & \cite{Niven2019} \\
 & Adversarial Evaluation & \cite{Niven2019} \\
\hline
ARC & Stress-test & \cite{Richardson2019} \\
\hline
OBQA & Stress-test & \cite{Richardson2019} \\
\hline
CoQA & Partial Baselines & \cite{Sugawara2019} \\
 & Manual Analyses & \cite{Yatskar2019} \\
\hline
DuoRC & Partial Baselines & \cite{Sugawara2019} \\
 & Stress-test & \cite{Dua2019a} \\
\hline
MCTest & Partial Baselines & \cite{Sugawara2019,Si2019} \\
 & Adversarial Evaluation & \cite{Si2019} \\
\hline
RACE & Partial Baselines & \cite{Sugawara2019,Si2019} \\
 & Adversarial Evaluation & \cite{Si2019} \\
\hline
SQuAD 2.0 & Partial Baselines & \cite{Sugawara2019} \\
 & Stress-test & \cite{Dua2019a} \\
 & Manual Analyses & \cite{Yatskar2019} \\
\hline
SWAG & Partial Baselines & \cite{Sugawara2019,Trichelair2018} \\
 & Adversarial Evaluation & \cite{Zellers2019,Zellers2018} \\
\hline
CNN & Manual Analyses & \cite{Chen2016} \\
 & Partial Baselines & \cite{Kaushik2019} \\
\hline
DailyMail & Manual Analyses & \cite{Chen2016} \\
\hline
DREAM & Partial Baselines & \cite{Si2019} \\
 & Adversarial Evaluation & \cite{Si2019} \\
\hline
MCScript & Partial Baselines & \cite{Si2019} \\
 & Adversarial Evaluation & \cite{Si2019} \\
\hline
MCScript 2.0 & Partial Baselines & \cite{Si2019} \\
 & Adversarial Evaluation & \cite{Si2019} \\
\hline
Hella\-SWAG & Adversarial Evaluation & \cite{Zellers2019} \\
\hline
ANLI & Adversarial Evaluation & \cite{Nie2019} \\
\hline
Narrative\-QA & Stress-test & \cite{Dua2019a} \\
\hline
Quoref & Stress-test & \cite{Dua2019a} \\
\hline
ROPES & Stress-test & \cite{Dua2019a} \\
\hline
WikiHop & Partial Baselines & \cite{chen2019understanding} \\
\hline
QNLI & Heuristics & \cite{Bras2020} \\
\hline
CBT & Partial Baselines & \cite{Kaushik2019} \\
 & Arch/Training Improvements & \cite{Grail} \\
\hline
Who-did-What & Partial Baselines & \cite{Kaushik2019} \\
\hline
bAbI & Partial Baselines & \cite{Kaushik2019} \\
\hline
SearchQA & Manual Analyses & \cite{Pugaliya2019} \\
\hline
ReClor & Heuristics & \cite{Yu2020} \\
\hline
Cambridge\-Dialogs & Arch/Training Improvements & \cite{Grail} \\
\hline
QuAC & Manual Analyses & \cite{Yatskar2019} \\
\hline
    \end{longtable}
    \label{tab:my_label}
%\end{table}
The following table shows those 36 datasets from Figure~\ref{fig:by-year} broken down by year, where no quantitative methods to describe possible spurious correlations have been applied yet:

\begin{longtable}{ p{.11\textwidth} | p{.81\textwidth} }
         \textbf{Year} & \textbf{Dataset} \\
         \hline

2015 & MedlineRTE \cite{Abacha2015}, WikiQA \cite{Yang2015}, DailyMail \cite{Hermann2015} \\
\hline
2016 & MSMarco \cite{Nguyen2016}, BookTest \cite{Bajgar2016}, SelQA \cite{Jurczyk2016}, WebQA \cite{Li2016} \\
\hline
2017 & SearchQA \cite{Dunn2017}, NewsQA \cite{Trischler2017}, GANNLI \cite{Starc2017}, TriviaQA \cite{Joshi2017}, CambridgeDialogs \cite{Wen2017} \\
\hline
2018 & PoiReviewQA \cite{Mai2018}, NarrativeQA \cite{Kocisky2018}, ReCoRd \cite{Zhang2018}, ARC \cite{Clark2018}, QuAC \cite{Choi2018}, emrQA \cite{Pampari2018}, ProPara \cite{Dalvi2018}, MedHop \cite{Welbl2018}, OBQA \cite{Mihaylov2018}, BioASQ \cite{Kamath2018} \\
\hline
2019 & BiPaR \cite{Jing2019}, NaturalQ \cite{Kwiatkowski2019}, ROPES \cite{Lin2019}, SherLIiC \cite{Schmitt2019}, CLUTRR \cite{Sinha2019}, PubMedQA \cite{Jin2019}, WIQA \cite{Tandon2019}, HELP \cite{Yanaka2019}, HEAD-QA \cite{Vilares2019}, CosmosQA \cite{Huang2019}, TWEET-QA \cite{Xiong2019}, RACE-C \cite{Liang2019}, VGNLI \cite{Mullenbach2019}, CEAC \cite{Liu2019} \\
\hline

\end{longtable}

%% file: appendix/inclusion-criteria.tex
\section{Inclusion Criteria for the Dataset Corpus}
\label{app:dataset-corpus}
We expand the collection of papers introducing datasets that were investigated or used by any publication in the original survey corpus (e.g. those shown in Figure~\ref{fig:by-dataset} by a Google Scholar search using the queries shown in Table~\ref{tab:queries}. We include a paper if it introduces a dataset for an NLI task according to our definition and the language of that dataset is English, otherwise we exclude it.

\begin{table}[h]
    \centering
    \begin{tabularx}{1\columnwidth}{X}
\texttt{allintitle: reasoning ("reading comprehension" OR "machine comprehension")   -image -visual -"knowledge graph" -"knowledge graphs"}\\
\texttt{allintitle: comprehension ((((set OR dataset) OR corpus) OR benchmark) OR "gold standard") -image -visual -"knowledge graph" -"knowledge graphs"}\\
\texttt{allintitle: entailment ((((set OR dataset) OR corpus) OR benchmark) OR "gold standard") -image -visual -"knowledge graph" -"knowledge graphs"}\\
\texttt{allintitle: reasoning ((((set OR dataset) OR corpus) OR benchmark) OR "gold standard") -image -visual -"knowledge graph" -"knowledge graphs"}\\
\texttt{allintitle: QA ((((set OR dataset) OR corpus) OR benchmark) OR "gold standard") -image -visual -"knowledge graph" -"knowledge graphs" -"open"} \\
\texttt{allintitle: NLI ((((set OR dataset) OR corpus) OR benchmark) OR "gold standard") -image -visual -"knowledge graph" -"knowledge graphs"}\\
\texttt{allintitle: language inference ((((set OR dataset) OR corpus) OR benchmark) OR "gold standard") -image -visual -"knowledge graph" -"knowledge graphs"} \\
\texttt{allintitle: "question answering" ((((set OR dataset) OR corpus) OR benchmark) OR "gold standard") -image -visual -"knowledge graph" -"knowledge graphs"} \\
    \end{tabularx}
    \caption{Google Scholar Queries for the extended dataset corpus}
    \label{tab:queries}
\end{table}